\documentclass[letterpaper, 10pt, conference]{IEEEtran}  %

\IEEEoverridecommandlockouts
\usepackage{float} 
\usepackage{amsmath,amssymb,amsfonts}
\usepackage{multirow}
\usepackage{algorithm}
\usepackage{algorithmic}
\usepackage{graphicx}
\usepackage{textcomp}
\usepackage{xcolor}
\usepackage{cite}
 \usepackage{booktabs} 
 \usepackage{threeparttable}  
 \usepackage{subcaption}
\def\BibTeX{{\rm B\kern-.05em{\sc i\kern-.025em b}\kern-.08em
    T\kern-.1667em\lower.7ex\hbox{E}\kern-.125emX}}

\title{Map-Agnostic And Interactive Safety-Critical Scenario Generation via Multi-Objective Tree Search}
\author{Wenyun Li$^{1,2}$, Zejian Deng$^{2}$ and Chen Sun$^{2,*}$
	\thanks{*This work was not supported by any organization}
    \thanks{$^{1}$ Department of Mathematics, The University of Hong Kong (HKU)}
	\thanks{$^{2}$ Department of Data and Systems Engineering, HKU}
    \thanks{*Corresponding author: Chen Sun \tt\small c87sun@hku.hk}%
}

\begin{document}

\maketitle
\begin{abstract}
Generating safety-critical scenarios is essential for validating the robustness of autonomous driving systems, yet existing methods often struggle to produce collisions that are both realistic and diverse while ensuring explicit interaction logic among traffic participants. This paper presents a novel framework for traffic-flow level safety-critical scenario generation via multi-objective Monte Carlo Tree Search (MCTS). We reframe trajectory feasibility and naturalistic behavior as optimization objectives within a unified evaluation function, enabling the discovery of diverse collision events without compromising realism. A hybrid Upper Confidence Bound (UCB) and Lower Confidence Bound (LCB) search strategy is introduced to balance exploratory efficiency with risk-averse decision-making. Furthermore, our method is map-agnostic and supports interactive scenario generation with each vehicle individually powered by SUMO's microscopic traffic models, enabling realistic agent behaviors in arbitrary geographic locations imported from OpenStreetMap. We validate our approach across four high-risk accident zones in Hong Kong's complex urban environments. Experimental results demonstrate that our framework achieves an 85\% collision failure rate while generating trajectories with superior feasibility and comfort metrics. The resulting scenarios exhibit greater complexity, as evidenced by increased vehicle mileage and CO\(_2\) emissions. Our work provides a principled solution for stress testing autonomous vehicles through the generation of realistic yet infrequent corner cases at traffic-flow level.
\end{abstract}

\section{Introduction}
In modern transportation systems, intelligent agents are increasingly assuming roles historically entrusted to human operators \cite{sun2023toward}. Irrespective of whether such systems are model-based or data-driven, their deployment necessitates rigorous safety validation \cite{ma2022verification}. Adaptive Stress Testing (AST) addresses this imperative by generating rare yet plausible extreme scenarios designed to probe the robustness and safety of a System Under Test (SUT)—typically an autonomous vehicle, robot, or alternative control policy. Accordingly, the systematic generation of realistic yet infrequent corner cases constitutes a pivotal endeavor in the safety assurance pipeline.

Existing methodologies have leveraged Reinforcement Learning (RL) to synthesize trajectories for stress testing, wherein reward functions incentivize agents to actively and efficiently traverse the state space in pursuit of trajectories that elicit failure or performance degradation in the SUT. Notwithstanding their efficacy, these approaches lack explicit objectives on states antecedent to the terminal condition, thereby compromising both the plausibility and naturalistic fidelity of the generated search outcomes. In contrast, the Multi-Objective Monte Carlo Tree Search (MCTS) framework proposed herein incorporates smoothness and feasibility metrics alongside rationality and naturalistic criteria as explicit constraints during trajectory optimization, thereby ensuring that resultant collision scenarios manifest both realism and plausibility. Furthermore, empirical evaluations substantiate that the traffic-flow level scenario generation methodology adopted in this work not only yields collision events but also produces scenarios of heightened complexity and emissions demands, thereby imposing more stringent requirements on the SUT's capacity to navigate intricate traffic conditions with augmented efficiency.

A fundamental limitation in data-driven scenario generation approaches resides in their propensity to execute scene modifications either devoid of physical plausibility and interactive dynamics or accompanied by only tenuous physical guarantees. Conversely, methodologies grounded in interactive dynamics offten suffer from insufficient scenario complexity—confined to simplistic highway configurations and limited agent counts—rendering them ill-equipped to accommodate the urban traffic flows confronting contemporary autonomous driving systems. Table~\ref{tab:feature_comparison} summarizes the feature comparison between our approach and prior studies, highlighting these gaps in previous work.  To surmount these shortcomings, the present study directly imports maps from Hong Kong's characteristically narrow and congested streetscape, wherein vehicles within the search space are instantiated as individual models powered by the SUMO (Simulation of Urban MObility) \cite{dlr127994} simulator. This configuration facilitates interactive search within densely interactive scenarios, engendering corner cases that manifest increasingly realistic and plausible collision events.

\begin{figure}[htbp]
\centering
\includegraphics[width=0.5\textwidth]{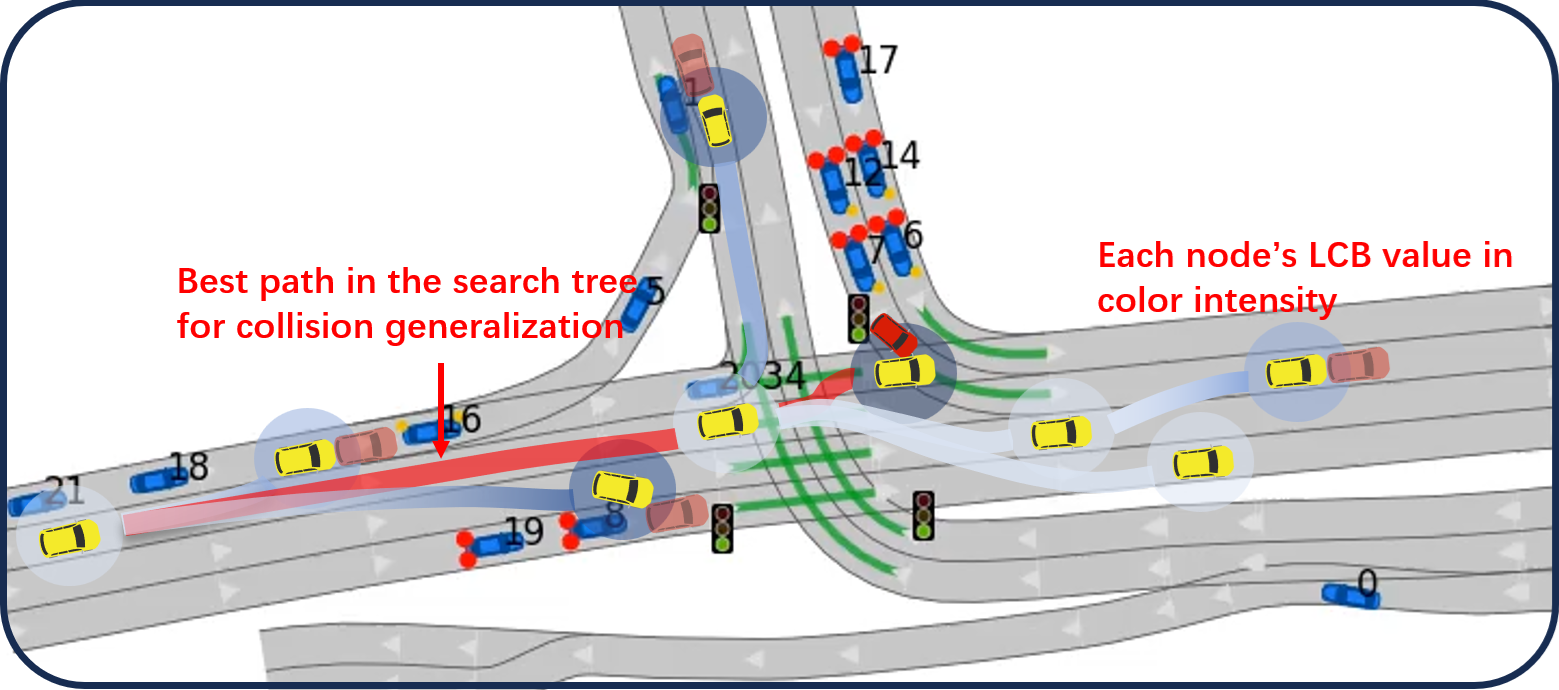}
\caption{Illustration of best path selection in tree search. The color intensity of each node represents its LCB value, with darker shades indicating higher (i.e., more favorable) lower confidence bounds. The red segments denote the selected best path, where the UCB-LCB strategy preferentially selects nodes with larger LCB values during final generation, thereby yielding more realistic collision scenarios.}
\label{fig:mcts_path_selection}
\end{figure}

Upper Confidence Bound (UCB) \cite{KocsisS06} constitutes a prevalent heuristic within MCTS, facilitating efficient exploration of the search space. However, within the specific context of collision scenario generation, considerations must extend beyond mere search efficiency to encompass the optimization of multiple objectives embedded within generated trajectories. To address this exigency, we employ Lower Confidence Bound (LCB) to effectuate risk-averse yet robust decision-making, selecting actions whose performance remains satisfactory even under pessimistic estimates, as illustrated in Figure~\ref{fig:mcts_path_selection}. This dual strategy judiciously balances the exploratory vigor of UCB with the decisional robustness afforded by LCB, while obviating the arduous task of tuning evaluation function weights encountered in MCTS frameworks.

The principal contributions of this work are threefold:

\begin{itemize}
    \item \textbf{Multi-Objective Trajectory Generation:} We introduce a multi-objective MCTS framework that explicitly enforces smoothness, feasibility, rationality, and naturalistic criteria during optimization, thereby generating traffic-flow level corner cases that are both physically plausible and behaviorally realistic.
    
    \item \textbf{Map-Agnostic Interactive Scenario Synthesis:} Our methodology accommodates arbitrary geographic inputs by directly importing maps from diverse regions—exemplified herein through Hong Kong's intricate streetscape—thereby demonstrating broad applicability. By leveraging SUMO-powered individual vehicle models within the search space, we enable interactive scenario exploration wherein generated corner cases emerge from logically consistent vehicle interactions.
    
    \item \textbf{Hybrid UCB-LCB Search Strategy:} We propose a hybrid search strategy that combines UCB's exploratory efficiency with LCB's decision-theoretic robustness. This integrated approach not only ensures efficient traversal of the search space but also enhances decision stability under uncertainty, while simultaneously circumventing the parameter-tuning overhead associated with conventional evaluation functions in MCTS.
\end{itemize}

The rest of the paper is organized as follows: Section~\ref{section.related} introduces the related work. Section~\ref{section.method} presents the proposed framework, detailing its core modules. Section~\ref{section.exp} describes the experimental setup, presents validation results, and provides in-depth analysis of the findings. Section~\ref{section.conclusion} concludes the paper with a summary of key contributions and discusses promising directions for future research.

\begin{table}[t]
\centering
\caption{Feature comparison with prior work}
\label{tab:feature_comparison}
\begin{threeparttable}
\begin{tabular}{l c c}
\toprule
\textbf{Method} & \textbf{Smooth/Feasible} & \textbf{Rational/Naturalistic} \\
\midrule
Ours & \checkmark & \checkmark \\
AdvSim~\cite{Wang_2021_CVPR} & \checkmark &  \\
Adaptive Search~\cite{adaptive2022failure} &  & \checkmark \\
Learning2Collide~\cite{learning2021collide} & \checkmark &  \\
ReGentS~\cite{yin2024regents} & \checkmark &  \\
AccidentSim~\cite{Zhang2025AccidentSimGV} & \checkmark & \checkmark \\
\bottomrule
\end{tabular}

\vspace{1ex}

\begin{tabular}{l c c c}  
\toprule
\textbf{Method} & \textbf{Interactive} & \textbf{Map Import} & \textbf{Scale} \\  
\midrule
Ours & \checkmark & \checkmark & Traffic Flow \\
AdvSim~\cite{Wang_2021_CVPR} &  &  & 1-5 Agents \\
Adaptive Search~\cite{adaptive2022failure} & \checkmark &  & 1-5 Agents \\
Learning2Collide~\cite{learning2021collide} &  &  & 2 Agents \\
ReGentS~\cite{yin2024regents} &  &  & Max 32 Agents \\
AccidentSim~\cite{Zhang2025AccidentSimGV} &  &  & 1-4 Agents \\
\bottomrule
\end{tabular}
\end{threeparttable}
\end{table}

\section{Related Work}\label{section.related}

Adaptive Stress Testing has emerged as a prominent framework for identifying safety-critical failures in autonomous systems across aerospace, driving, and maritime domains. Early work by Lee et al.~\cite{LeeMSGGSOK20} applied AST with Monte Carlo Tree Search to overcome partial observability in air collision avoidance systems. Subsequent research has extended this paradigm across diverse domains and fidelity levels. In autonomous driving, multiple studies have leveraged AST within simulation environments such as CARLA and Highway-Env to generate challenging scenarios for various controllers, ranging from PID and intelligent driver models to deep reinforcement learning policies~\cite{learning2021collide, finding2022failures, adaptive2022failure,Wang_2021_CVPR}. Methodological advancements include refined reward design that evolves from simple collision detection to considering longitudinal and lateral safety margins with respect to front, side, and rear vehicles~\cite{novel2023framework}, as well as the integration of domain knowledge and human-like reasoning into MCTS-based search processes~\cite{adaptive2022failure}. Beyond trajectory-level collision generation, recent work has extended safety-critical scenario synthesis to the video domain: Challenger~\cite{challenger2024} generates adversarial driving videos through iterative physics-aware trajectory optimization and rendering-compatible adversarial scoring; AccidentSim~\cite{Zhang2025AccidentSimGV} leverages real-world accident reports and fine-tuned LLMs to predict physically realistic post-collision trajectories; and SafeMVDrive~\cite{safemvdrive2024} enables multi-view safety-critical driving video synthesis in the real-world domain.

Beyond terrestrial driving, AST has demonstrated versatility in maritime autonomous systems and black-box language model planning. Recent studies have adapted AST for autonomous electric ferries, evolving from single-path velocity planners in Julia-based simulators to methodologies that generate more diverse and unexpected collision scenarios~\cite{green2024urban, evolutionary2024adaptive}. More recently, AST combined with MCTS has been employed to efficiently search the space of prompt perturbations for black-box large language model planners such as DeepSeek-R1 and Llama 3.2~\cite{chakraborty2026characterizingrobustnessblackboxllm}, underscoring the framework's adaptability to emerging AI architectures. Collectively, these studies illustrate a progression from finding any failure to discovering semantically meaningful, diverse, and multi-modal critical scenarios, motivating the need for adaptive risk-awareness strategies that balance exploration and exploitation.

\section{Methodology}\label{section.method}
The collision trajectory generation task is formalized within the framework of a Markov Decision Process (MDP), denoted by the tuple $\langle \mathcal{S}, \mathcal{A}, \mathcal{R}, \mathcal{P} \rangle$. Here, $\mathcal{S}$ represents the state space encompassing both traffic participants and the geographical environment, with $\mathbf{s}_t \in \mathcal{S}$ capturing the joint configuration of all dynamic agents and static scene elements at timestep $t$. $\mathcal{A}$ denotes the action space of traffic participants, where for agent $i$, the action $\mathbf{a}_t^{(i)} \in \mathcal{A}^{(i)}$ represents the control input applied at timestep $t$. The reward function (evaluation function in MCTS) $\mathcal{R}: \mathcal{S} \times \mathcal{A} \rightarrow \mathbb{R}$ assigns a scalar signal based on the state-action pair, while $\mathcal{P}: \mathcal{S} \times \mathcal{A} \times \mathcal{S} \rightarrow [0, 1]$ governs the state transition probabilities.

In the context of traffic scenario generation, a set of $N \in \mathbb{N}$ traffic participants $\mathcal{I} = \{1, 2, \ldots, N\}$ is considered, encompassing heterogeneous agent types—bicycles, cars, and trucks—each governed by a default behavior policy $\pi_{\text{default}}^{(i)}: \mathcal{S} \rightarrow \mathcal{A}^{(i)}$. A collision event involving $m \in \mathbb{N}_{\geq 2}$ participants is defined as a condition on the system state: a collision occurs at timestep $t$ if there exists a subset $\mathcal{C} \subseteq \mathcal{I}$ with $|\mathcal{C}| = m$ such that the spatial occupancy of any two distinct agents $j, k \in \mathcal{C}$ overlaps.

A specific participant $i \in \mathcal{C}$ is designated as the target agent whose action space is actively searched. Without loss of generality, at each timestep $t$, its action space $\mathcal{A}^{(1)}$ is discretized into a finite set of maneuver primitives, parameterized by steering angle $\theta_t \in \Theta \subset \mathbb{R}$ and acceleration rate $\alpha_t \in \mathrm{A} \subset \mathbb{R}$. The action executed by the target agent at timestep $t$ is therefore given by:

\begin{equation}
    \mathbf{a}_t^{(1)} = (\theta_t, \alpha_t) \in \mathcal{A}^{(1)} \subseteq \Theta \times \mathrm{A}
\end{equation}

The objective of collision trajectory generation is to search the target action space $\mathcal{A}^{(1)}$ for an optimal action sequence $\boldsymbol{\tau}^* = (\mathbf{a}_1^{(1)}, \mathbf{a}_2^{(1)}, \ldots, \mathbf{a}_T^{(1)})$—with other agents responding according to their respective policies $\pi_{\text{default}}^{(i)}$—that maximizes the cumulative discounted reward over a horizon of length $T$. Formally, this optimization problem is expressed as:

\begin{equation}
    \boldsymbol{\tau}^* = \arg\max_{\mathbf{a}_1^{(1)}, \ldots, \mathbf{a}_T^{(1)} \in \mathcal{A}^{(1)}} \mathbb{E}\left[ \sum_{t=1}^{T} \gamma^{t-1} \mathcal{R}(\mathbf{s}_t, \mathbf{a}_t^{(1)}) \right]
\end{equation}

where $\gamma \in [0, 1]$ is the discount factor, and the expectation is taken over the stochastic state transitions $\mathcal{P}$ and the default policies $\pi_{\text{default}}^{(i)}$ of non-target agents $i \in \mathcal{I} \setminus \{1\}$.

\textbf{Multi-Objective MCTS}
While Constrained Markov Decision Processes (CMDPs) constitute a well-established research domain, the specific instantiation of constrained Monte Carlo Tree Search has received comparatively limited attention in the literature~\cite{CMCTS}. The fundamental challenge stems from the nature of hard constraints: violations render solutions infeasible, necessitating the application of Lagrange multipliers to transform the constrained problem into an unconstrained formulation. This transformation typically involves recasting the original CMDP as a linear programming problem and subsequently solving its dual form—an undertaking that becomes computationally prohibitive within the high-fidelity simulation contexts considered herein.

Consequently, the present work adopts a principled departure from the constrained paradigm: rather than treating trajectory generation criteria such as feasibility and naturalistic behavior as hard constraints, we reframe them as optimization objectives. This design choice is motivated by the overarching objective of collision trajectory generation—namely, to induce system failures. Given that the very purpose is to produce failure cases, the imposition of hard constraints on the generated trajectories would be antithetical to the task at hand. Instead, multiple objectives are consolidated within a unified evaluation function, which is subsequently balanced through a Hybrid UCB-LCB Search Strategy that navigates trade-offs among competing criteria.

The resulting MCTS evaluation function takes the following form:

\begin{equation}
    \mathcal{R}(\mathbf{s}_t, \mathbf{a}_t^{(1)}) = \mathcal{R}_{\text{collision}} + \mathcal{C}_{\text{rationality}} + \mathcal{C}_{\text{naturalistic}}
\end{equation}

where each component is defined as follows:

\begin{itemize}
    \item $\mathcal{R}_{\text{collision}}$ denotes the collision reward, which incorporates a \textit{diversity bonus} designed to encourage the generation of distinct collision scenarios, thereby mitigating mode collapse and promoting exploration of the failure space.
    
    \item $\mathcal{C}_{\text{rationality}}$ represents a cost function quantifying the kinematic comfort of the generated trajectory. This term penalizes excessive longitudinal and lateral acceleration, substantial lateral offset from the lane center, and abrupt steering maneuvers (sharp turns).
    
    \item $\mathcal{C}_{\text{naturalistic}}$ constitutes a cost function, defined as a negative value, that encodes trajectory realism and safety considerations. It is operationalized through established safety metrics including Time-to-Collision (TTC) and Deceleration Rate to Avoid a Collision (DRAC) \cite{dlr127994}.
\end{itemize}

\textbf{Hybrid UCB-LCB Search Strategy}
The Upper Confidence Bound  algorithm addresses the exploration-exploitation dilemma by augmenting empirical reward estimates with an uncertainty bonus that encourages exploration of less-visited actions. This principle extends naturally to Monte Carlo Tree Search, where UCB serves as the tree policy—commonly referred to as UCT—guiding node selection during traversal. At a given node representing state $\mathbf{s}_t$, the algorithm selects the child node $\mathbf{a}_t^{(1)} $ that maximizes:
\begin{equation}\label{eq:ucb}
    \mathbf{a}_t^{(1)} = \arg\max_{\mathbf{a}_t^{(1)} \in \mathcal{A}^{(1)}} \left[ Q(\mathbf{s}_t,\mathbf{a}_t^{(1)}) + C \sqrt{\frac{\ln N(\mathbf{s}_t)}{N(\mathbf{s}_t,\mathbf{a}_t^{(1)})}} \right]
\end{equation}
where $Q(\mathbf{s}_t,\mathbf{a}_t^{(1)})$ is the current value estimate, $N(\mathbf{s}_t)$ the visitation count of the parent node, $C$ is an exploration constant, and $N(\mathbf{s}_t,\mathbf{a}_t^{(1)})$ the visitation count of the child node. This formulation guarantees convergence to optimal values as computational budget increases.

While UCB prioritizes exploration, the Lower Confidence Bound adopts a risk-averse counterpart by subtracting the uncertainty term, thereby selecting actions with consistently strong performance under pessimistic estimates. Following the notation in Eq.~\ref{eq:ucb}, the LCB criterion is defined as:
\begin{equation}\label{eq:lcb}
LCB(\mathbf{a}_t^{(1)})=Q(\mathbf{s}_t,\mathbf{a}_t^{(1)}) - C \sqrt{\frac{\ln N(\mathbf{s}_t)}{N(\mathbf{s}_t,\mathbf{a}_t^{(1)})}}
\end{equation}
This property makes LCB particularly suitable for safety-critical applications where worst-case guarantees are desirable \cite{KocsisS06}.

In our hybrid approach, we leverage both UCB and LCB to achieve efficient exploration while maintaining robust decision-making under uncertainty, as detailed in Algorithm~\ref{alg:hybrid-mcts}. The selection strategy switches between UCB and LCB based on two conditions: (i) when the tree depth ratio exceeds a threshold, or (ii) when the visitation counts of the current node and all its children surpass a minimum threshold. Under these conditions, the algorithm employs LCB for risk-averse decisions during both tree traversal and final action selection; otherwise, it defaults to UCB for efficient exploration.

\begin{algorithm}[t]
\caption{Hybrid UCB-LCB MCTS}
\label{alg:hybrid-mcts}
\begin{algorithmic}[1]
\REQUIRE Root state $\mathbf{s}_0$, max iterations $N_{\max}$, default vehicle policy $\pi_{\text{default}}^{(i)}$
\ENSURE Optimal action sequence $\boldsymbol{\tau}^*$

\STATE Initialize root node with current state

\FOR{$iter \gets 1$ to $N_{\max}$}
    \STATE \textbf{Selection:} Traverse tree from root:
    \STATE \hspace{\algorithmicindent} If unvisited child exists, select it
    \STATE \hspace{\algorithmicindent} Else if use\_LCB condition holds, select $\arg\max_{\mathbf{a}_t^{(1)}} \mathrm{LCB}(\mathbf{a}_t^{(1)})$
    \STATE \hspace{\algorithmicindent} Else select action according to Eq.~\ref{eq:ucb}
    \STATE \hspace{\algorithmicindent} Repeat until expandable node or max depth
    
    \STATE \textbf{Expansion:} If node not fully expanded, add new child for untried action
    
    \STATE \textbf{Simulation:} Execute rollout in SUMO with default vehicle policy $\pi_{\text{default}}^{(i)}$ to obtain cumulative reward $\mathcal{R}(\mathbf{s}_t, \mathbf{a}_t^{(1)})$
    
    \STATE \textbf{Backpropagation:} Update visitation counts and values along traversed path
\ENDFOR

\STATE \textbf{Return:} Optimal root action — if use\_LCB holds, select $\arg\max_{\mathbf{a}_t^{(1)}} \mathrm{LCB}(\mathbf{a}_t^{(1)})$; otherwise select most visited action
\end{algorithmic}
\end{algorithm}

\section{Experiments}\label{section.exp}
This section presents empirical evaluations of our proposed  framework for generating realistic yet infrequent collision corner cases in complex Hong Kong urban environments. We first describe the experimental setup, followed by quantitative comparisons with baseline methods across multiple trajectory quality metrics and scenario complexity indicators.

\subsection{Experimental Setup}

Our MCTS implementation leverages the simulation capabilities of SUMO, a microscopic and continuous traffic simulation package designed to handle large-scale networks \cite{dlr89233}. SUMO provides essential functionality for storing and backtracking simulation states—a prerequisite for tree-based search—along with built-in car-following and lane-changing models that enable interactive scenario exploration. Specifically, we employ the Extended Intelligent Driver Model (EIDM) \cite{Salles2022ExtendingTI} as the default behavior policy for non-target agents, with standard parameters governing reaction time, desired minimum time headway, and deceleration characteristics.

\begin{figure}
    \centering
    \includegraphics[width=0.95\linewidth]{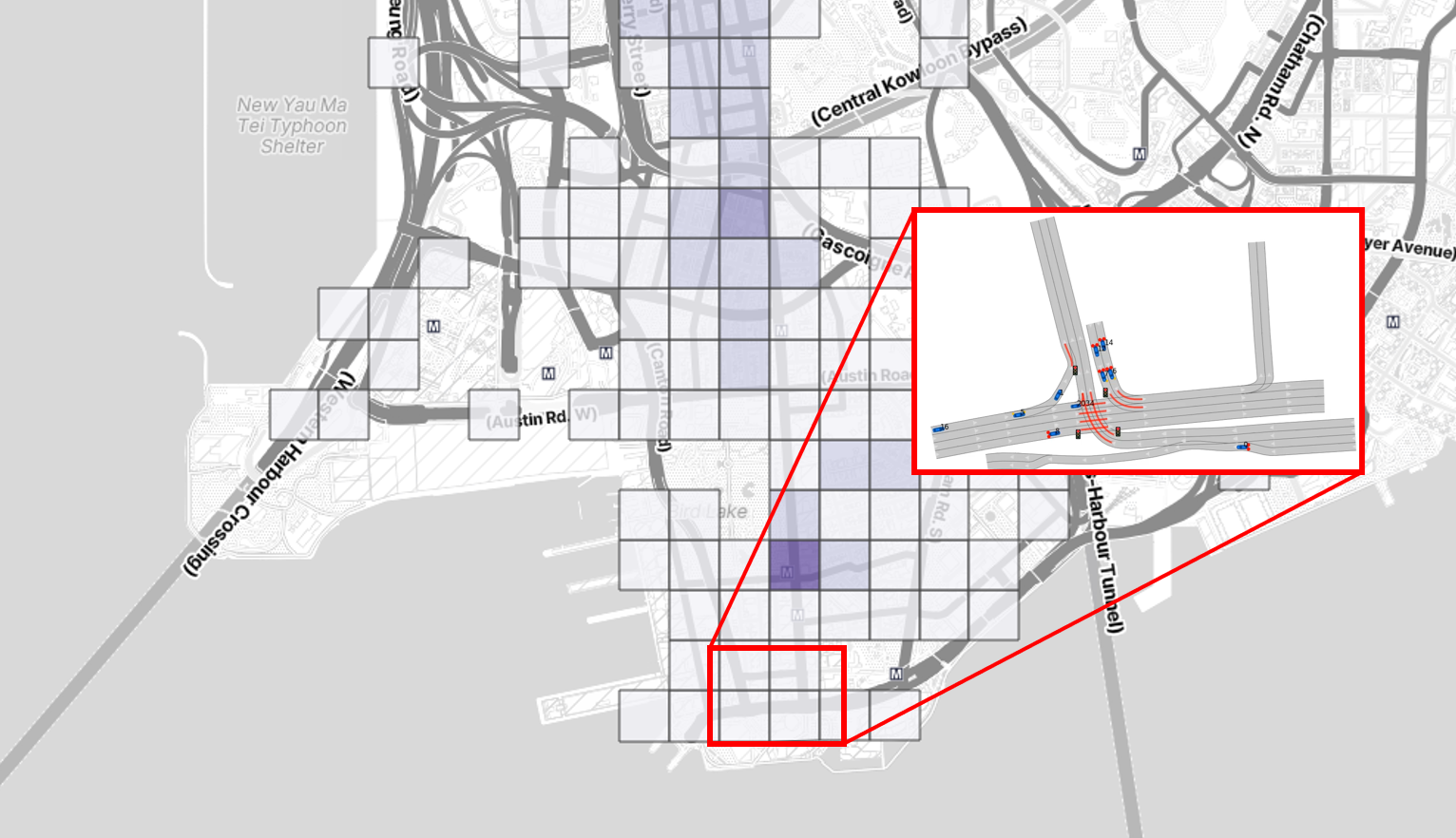}
    \caption{Illustration of selected Hong Kong street in Tsim Sha Tsui and traffic collision statistics. The geographic data and collision records are sourced from the Hong Kong Traffic Injury Collision Database \cite{hktrafficcollisions}.}
    \label{fig:TST}
\end{figure}

A key advantage of our framework lies in its map-agnostic design: scenes can be imported from arbitrary geographic locations using OpenStreetMap or CommonRoad \cite{7995802} formats. For validation, we select four high-risk accident zones across Hong Kong Island and Kowloon, encompassing diverse traffic patterns and road geometries as shown in Figure~\ref{fig:TST}. For each imported map, we utilize SUMO's official API to generate random traffic flows, followed by a stochastic number of warm-up simulation steps to establish realistic initial conditions.

\subsection{Evaluation of Collision Generation Capability}

We first assess the efficacy of our approach in generating collision events through 2000 runs in SUMO-based urban traffic environments. As reported in Table~\ref{tab:quantitative_comparison}, our Multi-Objective MCTS framework achieves a failure rate of 85\%, meaning that collision events are successfully induced in 85\% of search episodes. Moreover, each run yields an average of 428.33 novel collision trajectories—substantially higher than the variant without multi-objective constraints (381.17). This demonstrates that incorporating multi-objective criteria does not compromise collision generation capability; rather, it enhances the diversity of discovered failure modes. Note that when the target agent follows the default EIDM policy without active search, no collisions are observed.

\begin{table}[htbp]
\centering
\caption{Quantitative comparison on collision generation capability}
\label{tab:quantitative_comparison}
\resizebox{\columnwidth}{!}{%
\begin{tabular}{l c c c}
\toprule
\textbf{Metric} & \textbf{No Multi-Obj} & \textbf{Multi-Obj (Ours)} & \textbf{EIDM} \\
\midrule
Failure Rate (\%) & 75.60 & \textbf{85.00} & 0.00 \\
Collision Diversity & 381.17 & \textbf{428.33} & 0.00 \\
\bottomrule
\end{tabular}
}
\end{table}

\begin{figure}[htbp]
\centering
\begin{subfigure}[t]{0.24\textwidth}
\centering
\includegraphics[width=\textwidth]{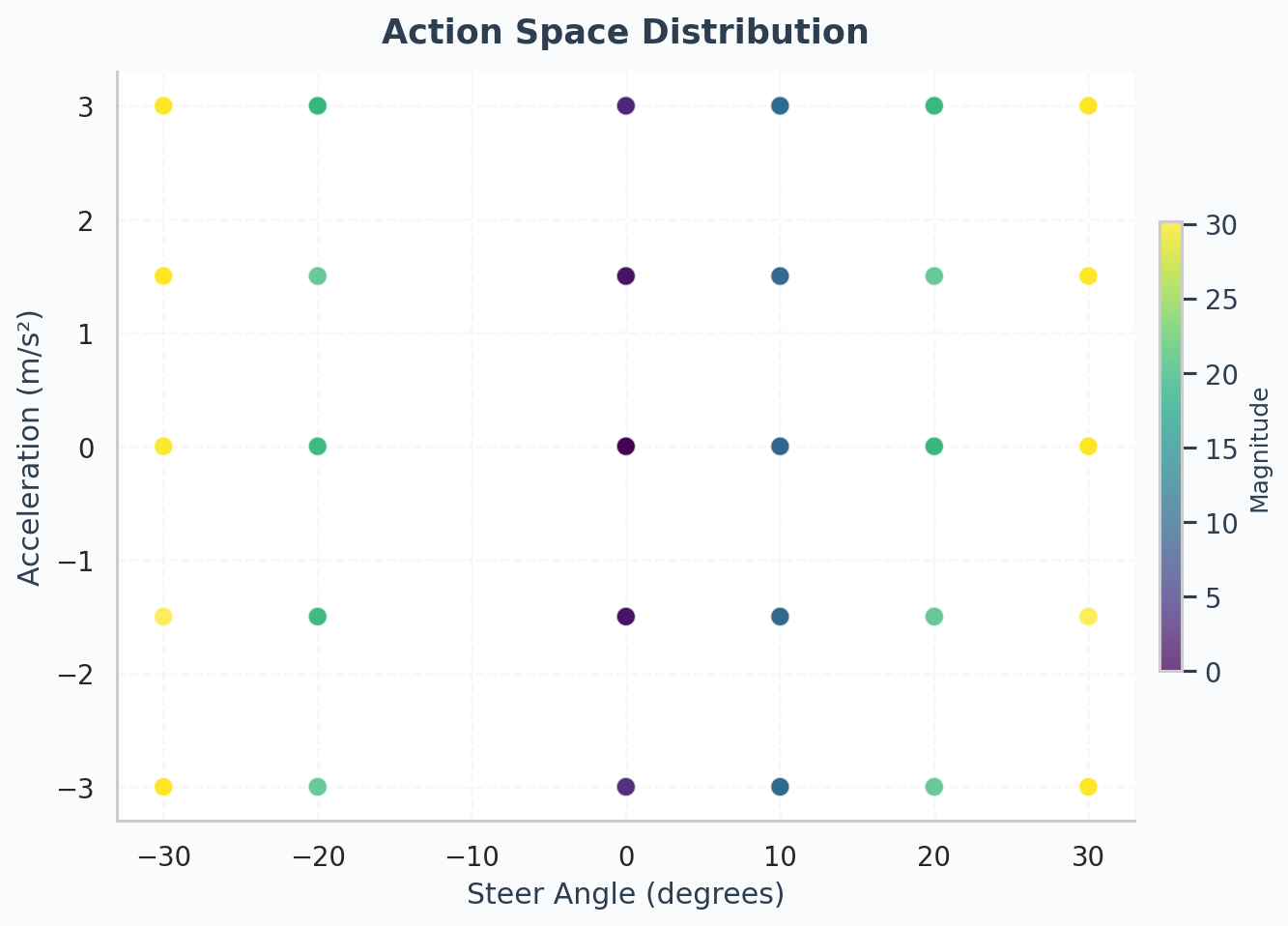}
\caption{Without multi-objective constraints}
\label{fig:first}
\end{subfigure}
\hfill
\begin{subfigure}[t]{0.24\textwidth}
\centering
\includegraphics[width=\textwidth]{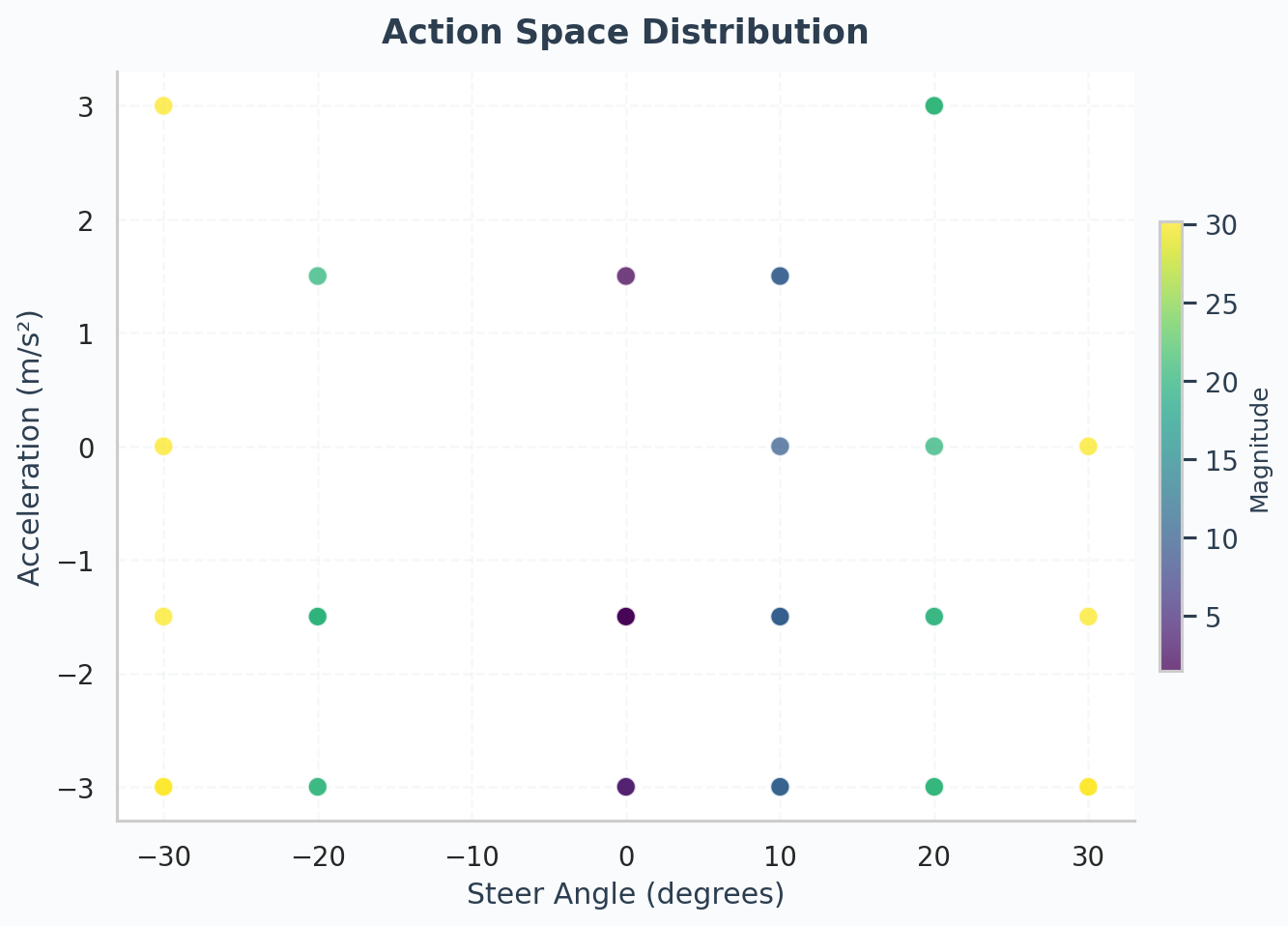}
\caption{With multi-objective constraints}
\label{fig:second}
\end{subfigure}
\caption{Scatter distribution of the action space}
\label{fig:comparison}
\end{figure}

Figure~\ref{fig:comparison} shows comparison of action space distributions during collision trajectory generation. The scatter plots visualize the distribution of discrete actions in the steering angle--acceleration plane. Color intensity indicates the relative frequency (magnitude) of each action being selected during the search process. Without feasibility and rationality constraints, the action distribution in Figure~\ref{fig:first} exhibits near-uniform coverage across dangerous maneuvers, lacking discernible preference patterns. While with multi-objective constraints incorporating feasibility and naturalistic criteria, the distribution concentrates predominantly on hard braking maneuvers, aligning with real-world collision causation where emergency deceleration is the predominant response, whereas acceleration toward other vehicles occurs less frequently and yields less realistic collision scenarios.

\subsection{Trajectory Quality and Scenario Complexity}

To comprehensively evaluate the generated scenarios, we employ a suite of metrics characterizing trajectory quality and scenario complexity, as shown in Table~\ref{tab:trajectory_metrics}. We compare three configurations: (i) multi-objective MCTS with standard UCB, (ii) our proposed hybrid UCB-LCB approach, and (iii) the default EIDM policy in 2000 runs. Longitudinal acceleration and Emergency Brake quantify the intended intensity of longitudinal maneuvers from the control/action perspective, indicating whether acceleration or deceleration events are excessively aggressive. Speed Change, derived from actual velocity variations in simulation, captures the realized acceleration intensity and reflects whether the executed maneuvers are overly abrupt. Sharp Turn penalizes excessive steering curvature, and Lateral Offset measures deviation from lane center, together assessing path-following fidelity.

The results reveal a clear advantage of our hybrid UCB-LCB strategy. While both UCB-based and UCB-LCB methods successfully discover collision trajectories from the search tree, the risk-averse decision-making facilitated by LCB yields collision scenarios that substantially outperform the purely exploratory UCB counterpart across metrics reflecting feasibility and comfort—where larger values indicate more favorable outcomes. Notably, in safety-related indicators such as TTC and DRAC, our method demonstrates a propensity to avoid overtly irrational collision trajectories that merely force contact regardless of realism, instead generating corner cases that remain within acceptable bounds of driving safety requirements.

From a complementary perspective, Distance—defined as the total mileage accumulated by all traffic participants across scenarios—and CO\(_2\) emissions (aggregated in grams) serve as proxies for scenario complexity and the operational demands imposed on the SUT. Our method achieves superior performance on both metrics compared to baselines, indicating that the generated scenarios are not only more numerous in collision events but also substantially more complex and challenging. This heightened complexity translates to more rigorous stress testing, pushing the SUT to operate at the boundaries of its capabilities.

\begin{figure}[htbp]
\centering
\begin{subfigure}[t]{0.48\textwidth}
\centering
\includegraphics[width=\textwidth]{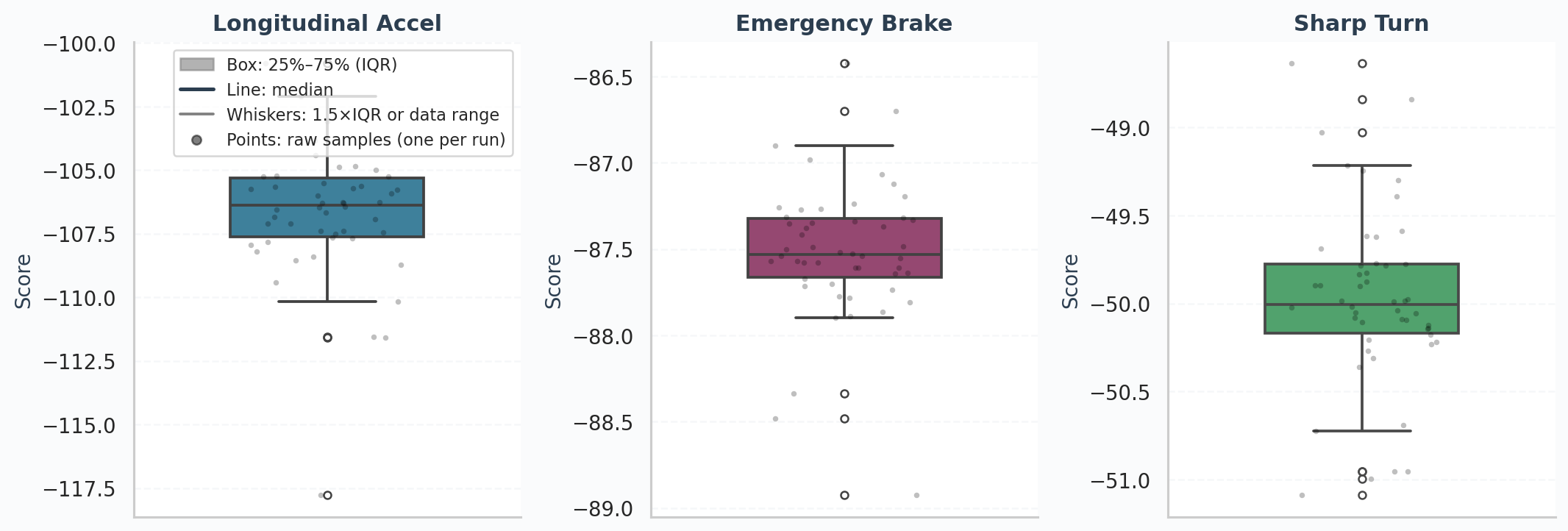}
\caption{UCB-only method}
\label{fig:ucb_boxplots}
\end{subfigure}
\hfill
\begin{subfigure}[t]{0.48\textwidth}
\centering
\includegraphics[width=\textwidth]{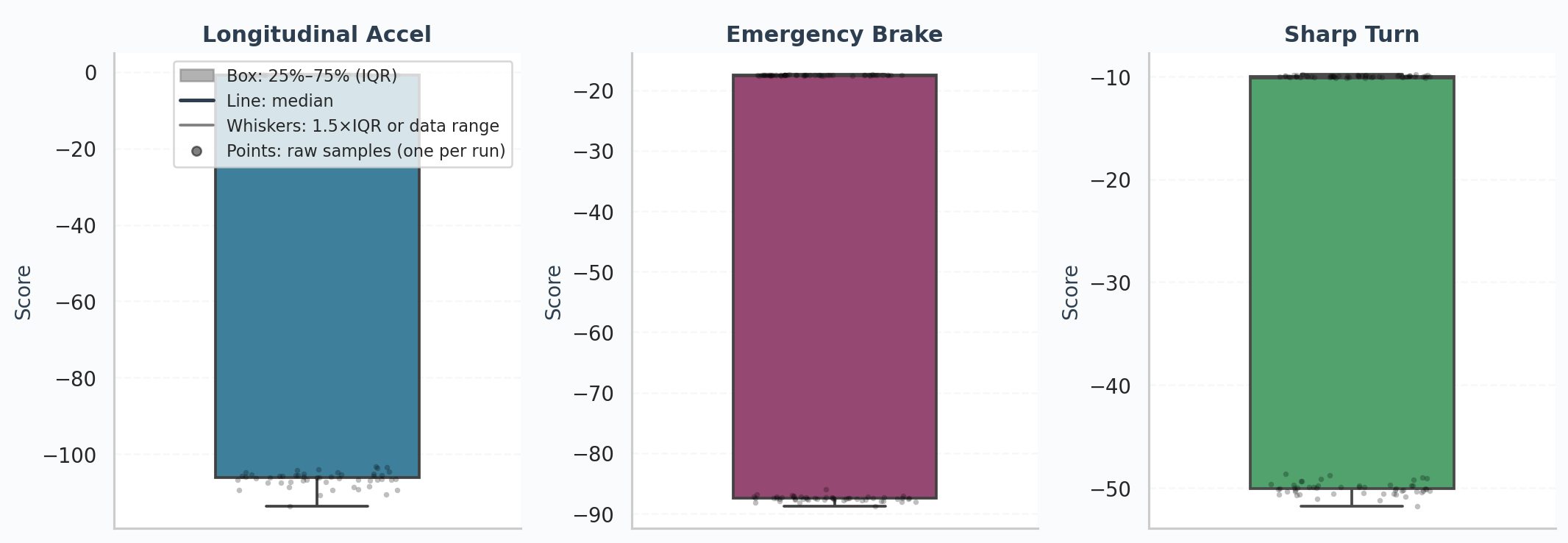}
\caption{Hybrid UCB-LCB method}
\label{fig:ucb_lcb_boxplots}
\end{subfigure}

\vspace{0.5cm}
\caption{Comparison of trajectory quality metrics between search strategies. }
\label{fig:boxplots_comparison}
\end{figure}

The boxplots in Figure~\ref{fig:boxplots_comparison} illustrate the distribution of longitudinal acceleration (m/s\(^2\)), emergency brake magnitude (m/s\(^2\)), and sharp turn curvature (rad/m) across generated collision scenarios, with boxplot elements defined in the legend. The comparison demonstrates how risk-averse LCB integration enhances trajectory feasibility and comfort: metrics generated by the hybrid UCB-LCB method exhibit more concentrated distributions around higher (i.e., more favorable) values, whereas those produced by the UCB-only approach display higher magnitudes and greater dispersion, indicating more aggressive and less controlled maneuvers.

\begin{table}[htbp]
\centering
\caption{Quantitative comparison on trajectory quality metrics}
\label{tab:trajectory_metrics}
\begin{tabular}{l c c c}
\toprule
\textbf{Metric} & \textbf{UCB} & \textbf{UCB-LCB (Ours)} & \textbf{EIDM} \\
\midrule
Longitudinal Accel & -106.66 & -51.63 & -7.46 \\
Lateral Accel & -12.01 & -4.60 & 0.00 \\
Emergency Brake & -87.51 & -51.17 & -100.00 \\
Sharp Turn & -49.97 & -29.26 & 0.00 \\
Speed Change & -17.50 & -12.08 & -12.00 \\
Lateral Offset & 0.22 & 0.62 & 1.00 \\
TTC & -1.53 & -0.58 & -1.04 \\
DRAC & -632.45 & -438.88 & 0.00 \\
Distance  & 1851.39 & 1885.67 & 1844.80 \\
Emission (CO2, g) & 33747.87 & 29923.47 & 26242.29 \\
\bottomrule
\end{tabular}
\end{table}

\section{Conclusion}\label{section.conclusion}

This paper has introduced a novel framework for traffic-flow level safety-critical scenario generation that addresses key limitations in existing stress testing methodologies. By reformulating trajectory feasibility and naturalistic behavior as optimization objectives within a multi-objective MCTS framework, we demonstrate that collision generation capability need not be sacrificed for realism. The proposed hybrid UCB-LCB search strategy represents a principled solution to the exploration-exploitation dilemma in safety-critical contexts: UCB ensures efficient traversal of the search space during early stages, while LCB enforces risk-averse decision-making near critical decision points, yielding trajectories that are both diverse and plausible.

A distinguishing feature of our approach lies in its support for interactive scenario search, where each vehicle is individually powered by SUMO's microscopic traffic models. This enables explicit interaction logic among traffic participants during the search process, ensuring that the resulting corner cases emerge from realistic agent behaviors rather than scripted or decoupled maneuvers. Coupled with our map-agnostic design that imports arbitrary geographic locations, the framework generates interactive safety-critical scenarios that reflect the complexities of real-world traffic dynamics.

Our empirical evaluations across four high-risk accident zones in Hong Kong's complex urban environments validate the efficacy of this approach. The framework achieves an 85\% collision rate with 428.33 novel collisions per run, substantially outperforming baselines. Generated trajectories exhibit superior feasibility and comfort, with increased mileage and CO\(_2\) emissions reflecting heightened scenario complexity.
Future work may extend to incorporate learning-based policies as default behaviors, and ultimately close the loop by deploying agents trained on generated datasets back into the scenario generation process. As autonomous driving systems continue to mature, such interactive stress testing methodologies will remain indispensable for ensuring their safe deployment in the unpredictable complexity of real-world traffic.
\bibliographystyle{IEEEtran}
\bibliography{ref}

\end{document}